\documentclass{article}

\PassOptionsToPackage{numbers, compress}{natbib}
\usepackage[final]{arxiv}

\usepackage{graphicx}
\graphicspath{ {figures/} }

\usepackage[utf8]{inputenc} % allow utf-8 input
\usepackage[T1]{fontenc}    % use 8-bit T1 fonts
\usepackage{hyperref}       % hyperlinks
\usepackage{url}            % simple URL typesetting
\usepackage{booktabs}       % professional-quality tables
\usepackage{amsfonts}       % blackboard math symbols
\usepackage{nicefrac}       % compact symbols for 1/2, etc.
\usepackage{microtype}      % microtypography

\usepackage{amsmath}
\usepackage{amssymb}
\usepackage{enumitem}
\usepackage{tablefootnote}
\usepackage[bottom]{footmisc}

\usepackage{wrapfig}
\usepackage{algorithmic}
\usepackage{subcaption}
\usepackage[ruled]{algorithm2e}

\usepackage{natbib}

\usepackage{tkz-euclide}
\usetkzobj{all}

\newcommand{\remove}[1]{{}}

\title{Scalable Methods for 8-bit Training of Neural Networks}

\author{
  Ron Banner$^{1}$\thanks{Equal contribution}, \hfill Itay Hubara$^{2}$\footnotemark[1],\hfill Elad Hoffer$^{2}$\footnotemark[1],  \hfill Daniel Soudry$^{2}$\\
  \texttt{\{itayhubara, elad.hoffer, daniel.soudry\}@gmail.com} \\
  \texttt{\{ron.banner\}@intel.com} \\
}

\begin{document}

\maketitle
(1) Intel - Artificial Intelligence Products Group (AIPG)\\
(2) Technion - Israel Institute of Technology, Haifa, Israel\\

\begin{abstract}
Quantized Neural Networks (QNNs) are often used to improve network efficiency during the inference phase, i.e. after the network has been trained. Extensive research in the field suggests many different quantization schemes. Still, the number of bits required, as well as the best quantization scheme, are yet unknown. Our theoretical analysis suggests that most of the training process is robust to substantial precision reduction, and points to only a few specific operations that require higher precision.  Armed with this knowledge, we quantize the model parameters,  activations and layer gradients to 8-bit, leaving at a higher precision only the final step in the computation of the weight gradients. Additionally, as QNNs require batch-normalization to be trained at high precision, we introduce Range Batch-Normalization (BN) which has significantly higher tolerance to quantization noise and improved computational complexity. Our simulations show that Range BN is equivalent to the traditional batch norm if a precise scale adjustment, which can be approximated analytically, is applied. To the best of the authors' knowledge, this work is the first to quantize the weights, activations, as well as a substantial volume of the gradients stream, in all layers (including batch normalization) to 8-bit while showing state-of-the-art results over the ImageNet-1K dataset.

% We show there is no inherent difficulty in training QNNs using "Quantized Back-Propagation" (QBP), and that with 8-bit we can achieve state of the art results over ImageNet-1K dataset. Using QBP on dedicated hardware has the potential to significantly improve the execution efficiency (\emph{e.g.}, reduce dynamic memory footprint, memory bandwidth, and computational energy) and speed up the training process with appropriate hardware support.
\end{abstract}

\section{Introduction}

Deep Neural Networks (DNNs) achieved remarkable results in many fields making them the most common off-the-shelf approach for a wide variety of machine learning applications. However, as networks get deeper, using neural network (NN) algorithms and training them on conventional general-purpose digital hardware is highly inefficient. The main computational effort is due to massive amounts of multiply-accumulate operations (MACs) required to compute the weighted sums of the neurons’ inputs and the parameters' gradients.

Much work has been done to reduce the size of networks. The conventional approach is to compress a trained (full precision) network \cite{chen2015compressing,ullrich2017soft,jaderberg2014speeding} using weights sharing, low rank approximation, quantization, pruning or some combination thereof. For example, Han et al., 2015 \cite{han2015deep} successfully pruned several state-of-the-art large-scale networks and showed that the number of parameters can be reduced by an order of magnitude.

Since training neural networks requires approximately three times more computation power than just evaluating them, quantizing the gradients is a critical step towards faster training machines. Previous work demonstrated that by quantizing network parameters and intermediate activations during the training phase more computationally efficient DNNs could be constructed. 

Researchers \citep{gupta2015deep,das2018mixed} have shown that 16-bit is sufficient precision for most network training but further quantization (i.e., 8-bit) results with severe degradation. Our work is the first to almost exclusively train at 8-bit without harming classification accuracy. This is addressed by overcoming two main obstacles known to hamper numerical stability: batch normalization and gradient computations.

The traditional batch normalization \cite{ioffe2015batch} implementation requires the computation of the sum of squares, square-root and reciprocal operations; these require high precision (to avoid zero variance) and a large dynamic range. It should come as no surprise that previous attempts to use low precision networks did not use batch normalization layers \cite{wu2018training} or kept them in full precision \cite{zhou2016dorefa}. This work replaces the batch norm operation with range batch-norm (range BN) that normalizes inputs by the range of the input distribution (i.e., $\max(x) - \min(x)$). This measure is more suitable for low-precision implementations. Range BN is shown analytically to approximate the original batch normalization by multiplying this range with a scale adjustment that depends on the size of the batch and equals to $ (2\cdot\ln(n))^{-0.5}$. Experiments on ImageNet with Res18 and Res50 showed no distinguishable difference between accuracy of Range BN and traditional BN. 

The second obstacle is related to the gradients quantization. Given an upstream gradient $g_l$ from layer $l$, layer $l-1$ needs to apply two different matrix multiplications: one for the layer gradient $g_{l-1}$ and the other for the weight gradient $g_W$ which are needed for the update rule. Our analysis indicates that the statistics of the gradient $g_l$ violates the assumptions at the crux of common quantization schemes. As such, quantizing these gradients constitutes the main cause of degradation in performance through training. Accordingly, we suggest to use two versions of layer gradients $g_l$, one with low-precision (8-bit) and another with higher-precision (16-bit). The idea is to keep all calculations with $g_l$ that does not involve a performance bottleneck at 16 bits, while the rest at 8 bits. As the gradients $g_W$ are required only for the weight update, they are computed using the 16 bits copy of $g_l$. On the other hand, the gradient $g_{l-1}$ is required for the entire backwards stream and as such it is computed using the corresponding 8-bit version of $g_l$. In most layers of the DNN these computations can be performed in parallel. Hence $g_W$ can be computed at high precision in parallel with $g_{l-1}$, without interrupting the propagation of $g_l$ to lower layers.  We denote the use of two different arithmetic precision operations in the differentiation process as "Gradients Bifurcation".

\section{Previous Work}
While several works \citep{gupta2015deep,das2018mixed} have shown that training at 16-bit is sufficient for most networks, more aggressive quantization schemes were also suggested \citep{zhou2016dorefa,miyashita2016convolutional,lin2017towards,hubara2016quantized}. In the extreme case, the quantization process used only one bit which resulted in binarized neural networks (BNNs) \cite{hubara2016binarized} where both weights and activations were constrained to -1 and 1. However, for more complex models and challenging datasets, the extreme compression rate resulted in a loss of accuracy. Recently, \citet{mishra2017wrpn} showed that this accuracy loss can be prevented by merely increasing the number of filter maps in each layer, thus suggesting that quantized neural networks (QNNs) do not possess an inherent convergence problem. Nevertheless, increasing the number of filter maps enlarge quadratically the number of parameters, which raises questions about the efficiency of this approach. 

In addition to the quantization of the forward pass, a growing interest is directed towards the quantization of the gradient propagation in neural networks. A fully quantized method, allowing both forward and backward low-precision operations will enable the use of dedicated hardware, with considerable computational, memory, and power benefits. Previous attempts to discretize the gradients managed to either reduce them to 16-bit without loss of accuracy \citep{das2018mixed} or apply a more aggressive approach and reduce the precision to 6-8 bit \cite{zhou2016dorefa,hubara2016quantized} with a noticeable degradation. Batch normalization is mentioned by  \cite{wu2018training} as a bottleneck for network quantization and is either replaced by a constant scaling layer kept in full precision, or avoided altogether; this clearly has some impact on performance (e.g., AlexNet trained over ImageNet resulted with top-1 error of $51.6\%$, where the state of the art is near $42\%$) and better ways to quantize normalization are explicitly called for. Recently L1 batch norm with only linear operations in both forward and backward propagation was suggested by \cite{wu2018l1,hoffer2018norm} with improved numerical stability. Yet, our experiments show that with 8-bit training even L1 batch norm is prone to overflows when summing over many large positive values. Finally, \citet{wen2017terngrad} focused on quantizing the gradient updates to ternary values to reduce the communication bandwidth in distributed systems.

We claim that although more aggressive quantization methods exist, 8-bit precision may prove to have a "sweet-spot" quality to it, by enabling training with no loss of accuracy and without modifying the original architecture. Moreover, we note that 8-bit quantization is better suited for future and even current hardware, many of which can already benefit from 8-bit operations \cite{intel_8bit}. 
So far, to the best of our knowledge, no work has succeeded to quantize the activations, weights, and gradient of all layers (including batch normalization) to 8-bit without any degradation.
% Recently \citet{anderson2017high} suggested an explanation to why extreme quantization (binarization) works. They hypothesize that Binarized neural Networks (BNNs) \cite{hubara2016binarized} work because of the high dimensional geometry of binary vectors. Specifically, they show that the angle between the continuous vectors and its binarized counterpart is relatively small ($\approx 37^{\circ}$) when the vector dimensions go to infinity. This study aims to generalize this work and thus set the foundation for understanding QNNs. In particular, we prove that:

% \begin{itemize}

% \item The expected angel for binarized vector is always less than 37 degrees.
% \item The angel for ternarized is the smallest when delta is equal to $0.6\sigma$ and equals to 25.8 degrees. 
% \item The expected cosine similarity between a tensor and its quantized  version  is derived theoretically and shown to be at most 99.2\% for 8-bit, 98.4\% for 7 bits, 97.1\% for 6 bits, 94.6\% for 5 bits, 90.5\% for 4 bits, 84.3\% for 3 bits and 77.1\% for 2  bits. : 
% \item The best clamping value for quantized tensor is given by 
% \end{itemize}

\section{Range Batch-Normalization}

 For a layer with $n\times d-$dimensional input $x=(x^{(1)},x^{(2)},...,x^{(d)})$,  traditional batch norm normalizes each dimension 
\begin{equation}
%\hat{x}^{(d)}=\sqrt{\frac{2}{\pi}}\cdot\frac{x^{(d)}-\mu^d]}{\sqrt{\mathrm{Var}[x^{(d)]}}
\label{L2_BN_equation}
\hat{x}^{(d)}=\frac{x^{(d)}-\mu^d}{ \sqrt{\mathrm{Var}[x^{(d)}]}},
\end{equation}
where  $\mu^d$ is the expectation over $x^{(d)}$,  $n$ is the batch size and $\mathrm{Var}[x^{(d)}]=\frac{1}{n}||{x^{(d)}-\mu^d}||_{2}^2$.
The term $\sqrt{\mathrm{Var}[x^{(d)}]}$ involves sums of squares that can lead to numerical instability as well as to arithmetic overflows when dealing with large values.  The Range BN method replaces the above term by normalizing according to the range of the input distribution (i.e., $\max(\cdot)- \min(\cdot)$), making it more tolerant to quantization. For a layer with $d-$dimensional input $x=(x^{(1)},x^{(2)},...,x^{(d)})$,  Range BN normalizes each dimension
\begin{equation}\label{range}
    \hat{x}^{(d)}=\frac{x^{(d)}-\mu^d}{C(n)\cdot\text{range}(x^{(d)}-\mu^d)},
\end{equation}
where $\mu^d$ is the expectation over $x^{(d)}$, $n$ is the batch size,  $C(n)=\frac{1}{\sqrt{2\cdot\ln(n)}}$  is a scale adjustment term, and $\text{range}(x) = \max(x)-\min(x)$. 

The main idea behind Range BN is to use the scale adjustment $C(n)$ to approximate the standard deviation $\sigma$ (traditionally being used in vanilla batch norm) by multiplying it with the range of the input values. Assuming the input follows a Gaussian distribution, the range (spread) of the input is highly correlated with the standard deviation magnitude.  Therefore by normalizing the range by $C(n)$ we can estimate $\sigma$. Note that the Gaussian assumption is a common approximation (e.g., \citet{soudry2014}), based on the fact that the neural input $x^{(d)}$ is a sum of many inputs, so we expect it to be approximately Gaussian from the central limit theorem.

We now turn to derive the normalization term  $C(n)$. The expectation of maximum of Gaussian random variables are bounded as follows \citep{kamath2015bounds}:

\begin{equation}\label{max}
\ 0.23\sigma\cdot\sqrt{\ln(n)}\leq E[\max({x^{(d)}-\mu^d})]\leq \sqrt{2}\sigma\sqrt{\ln(n).}
\end{equation}

Since ${x^{(d)}-\mu^d}$ is symmetrical with respect to zero (centred at zero and assumed gaussian), it holds that  $E[\max(\cdot)]=-E[\min(\cdot)]$; hence,

\begin{equation}\label{min}
\ 0.23\sigma\cdot\sqrt{\ln(n)} \leq -E[\min({x^{(d)}-\mu^d})]\leq \sqrt{2}\sigma\sqrt{\ln(n).}
\end{equation}

Therefore, by summing Equations \ref{max} and \ref{min} and multiplying the three parts of the inequality by the normalization term $C(n)$, Range BN in Eq. \ref{range} approximates the original standard deviation measure $\sigma$ as follows: 

$$ 0.325\sigma\leq C(n)\cdot\text{range}(x^{(d)}-\mu^d)\leq 2\cdot\sigma$$

Importantly, the scale adjustment term $C(n)$ plays a major role in  RangeBN success.  The performance was degraded in simulations when $C(n)$ was not used or modified to nearby values. 

\section{Quantized Back-Propagation} \label{sec:QBP}

\paragraph{Quantization methods:}
 Following \cite{wu2016google} we used the GEMMLOWP quantization scheme as decribed in Google's open source library \citep{Benoit2017gemmlowp}. A detailed explanation of this approach is given in Appendix \ref{app:GEMMLOWP}. While GEMMLOWP is widely used for deployment, to the best of the authors knowledge this is the first time GEMMLOWP quantization is applied for training. Note that the activations maximum and minimum values were computed by the range BN operator, thus finding the normalization scale as defined in Appendix \ref{app:GEMMLOWP} does not require additional $O(n)$ operations.

Finally we note that a good convergence was achieved only by using stochastic rounding \cite{gupta2015deep} for the gradient quantization. This behaviour is not surprising as the gradients will serve eventually for the weight update thus unbiased quantization scheme is required to avoid noise accumulation.

\paragraph{Gradients Bifurcation:}
In the back-propagation algorithm we recursively calculate the gradients of the loss function $\mathcal{L}$ with respect to $I_{\ell}$, the input of the $\ell$ neural layer, 
\begin{equation}
    g_{\ell} =\frac{\partial{\mathcal{L}}}{\partial{I_{\ell}}},   
\end{equation}
starting from the last layer. 
Each layer needs to derive two sets of gradients to perform the recursive update. The layer activation gradients:
\begin{equation} \label{eq:grad_layer}
     g_{\ell-1} = g_{\ell}W_{\ell}^T,
\end{equation}
served for the Back-Propagation (BP) phase thus passed to the next layer,and the weights gradients
\begin{equation}\label{eq:grad_weight}
     g_{W_{\ell}} = g_{\ell}I_{\ell-1}^T,
\end{equation}
used to updated the weights in  layer $\ell$. Since the backward pass requires twice the amount of multiplications compared to the forward pass, quantizing the gradients is a crucial step towards faster training machines. Since $g_{\ell}$, the gradients streaming from layer $\ell$, are required to compute $g_{\ell-1}$, it is important to expedite the matrix multiplication described in Eq.\ref{eq:grad_layer}. The second set of gradient derive in Eq.\ref{eq:grad_weight} is not required for this sequential process and thus we choose to keep this matrix multiplication in full precision. We argue that the extra time required for this matrix multiplication is comparably small to the time required to communicate the gradients $g_{\ell}$. Thus, in this work the gradients used for the weight gradients derivation are still in float. In section \ref{sec:experiments}, we show empirically that bifurcation of the gradients is crucial for high accuracy results.
\paragraph{Straight-Through Estimator:}
Similar to previous work \cite{hubara2016binarized,mishra2017wrpn}, we used the straight-through estimator (STE) approach to approximate differentiation through discrete variables. This is the most simple and hardware friendly approach to deal with the fact that the exact derivative of discrete variables is zero almost everywhere. 

\section{When is quantization of neural networks possible? }\label{theory}

This section provides some of the foundations needed for understanding the internal representation of quantized neural networks. It is well known that when batch norm is applied after a convolution layer, the output is invariant to the norm of the weight on the proceeding layer \cite{ioffe2015batch} i.e., $BN(C\cdot W \cdot x)=BN(W \cdot x)$ for any given constant $C$. This quantity is often described geometrically as the norm of the weight tensor, and in the presence of this invariance,  the only measure that needs to be preserved upon quantization is the \textit{directionality} of the weight tensor. In the following we show that quantization preserves the direction (angle) of high-dimensional vectors when $W$ follows a Gaussian distribution.

More specifically, for networks with $M$-bit fixed point representation, the angle is preserved when the number of quantization levels $2^M$ is much larger than $ \sqrt{2\ln(N)}$, where $N$ is the size of quantized vector. This shows that significant quantization is possible on practical settings. Taking for example the dimensionality of the joint product in a batch with 1024 examples corresponding to the last layer of ResNet-50, we need no more than 8-bit of precision to preserve the angle well (i.e., $ \sqrt{2\ln(3\cdot3\cdot2048\cdot1024)}=5.7 << 2^8$). We stress that this result heavily relays on values being  distributed according to a Gaussian distribution, and suggests why some vectors are robust to quantization (e.g., weights and activations) while others are more fragile (e.g., gradients).

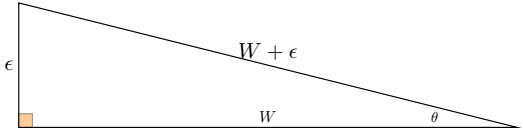
\begin{wrapfigure}{r}{0.6\textwidth}
%\begin{figure}[h] 

\resizebox {0.5\columnwidth} {!}{
\begin{tikzpicture}[thick]
\coordinate (O) at (0,3);
\coordinate (A) at (0,0);
\coordinate (B) at (12,0);
\draw (O)--(A)--(B)--cycle;
\tkzLabelSegment[left](O,A){\scalebox{1.5}{$\epsilon$}}
\tkzLabelSegment(O,B){\scalebox{1.5}{$W+\epsilon$}}
\tkzLabelSegment[above](A,B){\scalebox{1.1}{$W$}}

\tkzMarkRightAngle[fill=orange,size=0.33,opacity=.4](O,A,B)% square angle here
\tkzLabelAngle[pos = 2](O,B,A){$\theta$}

\end{tikzpicture}
}

\caption{ Graphic illustration of the angle between full precision vector $W$ and its low precision counterpart which we model as $W+\epsilon$ where $\epsilon\sim\mathbb{U}(-\Delta/2,\Delta/2)$}\label{fig:analysis}
%\end{figure}
\end{wrapfigure}

\pagebreak

\subsection{Problem Statement} \label{sec:Problem}

Given a vector of weights $W=(w_0,w_1,...,w_{N-1})$, where the weights follow a Gaussian distribution  $W \sim N(0,\sigma)$, we would like to measure the cosine similarity (i.e., cosine of the angle) between $W$ and $Q(W)$, where $Q(\cdot)$ is a quantization function. More formally, we are interested in estimating the following geometric measure:
\begin{equation}\label{eq:1}
\cos(\theta)=\dfrac{W\cdot Q(W)}{||W||_2\cdot ||Q(W)||_2}
\end{equation}

We next define the quantization function $Q(\cdot)$ using a fixed quantization step between adjacent quantified levels as follows:
\begin{equation}
Q(x) = \Delta\cdot\Big(\Big\lfloor\dfrac{x}{\Delta}\Big\rfloor + \dfrac{1}{2}\Big) \text{,  where       }                         \Delta= \dfrac{\max(|W|)}{2^{M}}
\end{equation}
We consider the case where quantization step $\Delta$ is much smaller than $\text{mean}(|W|)$. Under this assumption correlation between $W$ and quantization noise $W-Q(W) =(\epsilon_0,\epsilon_1,...,\epsilon_{N-1})$ is negligible, and can be approximated as an additive noise. Our model assumes an additive quantization noise $\Bar{\epsilon}$  with a uniform distribution i.e.,  $\epsilon_i \sim  \mathcal{U}[-\Delta/2,\Delta/2]$ for each index $i$. Our goal is to estimate the angle between $W$ and $W + \Bar{\epsilon}$ for high dimensions (i.e., $N \to \infty$).

\subsection{Angle preservation during quantization}

In order to estimate the angle between $W$ and $W+\epsilon$, we first estimate the angle between $W$ and $\epsilon$. It is well known that if $\epsilon$ and $W$ are independent, then at high dimension the angle between $W$ and $\epsilon$ tends to $\dfrac{\pi}{2}$ \cite{biau2015high} i.e., we get a right angle triangle with  $W$ and $\epsilon$ as the legs, while $W+\epsilon$ is the hypotenuse as illustrated in Figure \ref{fig:analysis}-right. The cosine of the angle $\theta$ in that triangle can be approximated as follows: 
\begin{equation}\label{eq:22}
\cos(\theta) =  \dfrac{||W||}{||W+\epsilon||}\geq \dfrac{||W||}{||W||+||\epsilon||}
\end{equation}

Since $W$ is Gaussian, we have that $E(||W||)\cong \sqrt{N}\sigma$ in high dimensions \cite{chandrasekaran2012convex}. Additionally, in Appendix \ref{app:N-bit} we show that  $E(||\Bar{\epsilon}||)\leq \sqrt{N/12}\cdot \Delta$. Moreover, at high dimensions, the relative error made as considering $E||X||$ instead of the random variable $||X||$ becomes asymptotically negligible \cite{biau2015high}. Therefore, the following holds in high dimensions:
\begin{equation}\label{eq:40}
\cos(\theta)  \geq \dfrac{\sigma}{\sigma+E(\Delta)/\sqrt{12}}= \dfrac{2^M\cdot\sigma}{2^M\cdot\sigma+ E(\max(|W|))/\sqrt{12}}
\end{equation}
Finally, $E(\max(W))\leq \sqrt{2}\sigma\sqrt{\ln(N})$ when $W$ follows a Gaussian distribution \citep{kamath2015bounds}, establishing the following: 
\begin{equation}\label{eq:41}
\cos(\theta) \geq \dfrac{2^{M}}{2^{M} + \sqrt{\ln N}/\sqrt{6}}
\end{equation}
Eq. \ref{eq:41} establishes that when $2^M >> \sqrt{\ln(N)}$ the angle is preserved during quantization. It is easy to see that in most practical settings this condition holds even for challenging quantizations. Moreover, this results highly depends on the assumption made about the Gaussian distribution of $W$ (transition from equation \ref{eq:40} to equation \ref{eq:41}).

\section{Experiments}\label{sec:experiments}

We evaluated the ideas of Range Batch-Norm and Quantized Back-Propagation on multiple different models and datasets. The code to replicate all of our experiments is available on-line \footnote{https://github.com/eladhoffer/quantized.pytorch}. 
%We begin by experiments done to gain further insights about the internal representations of quantized neural networks beyond the  analysis of section \ref{theory}. 

\subsection{Experiment results on cifar-10 dataset}

\begin{figure}[h] 
\vspace{-10pt}
\begin{subfigure}{.45\textwidth}
\centerline{\includegraphics[width=\columnwidth]{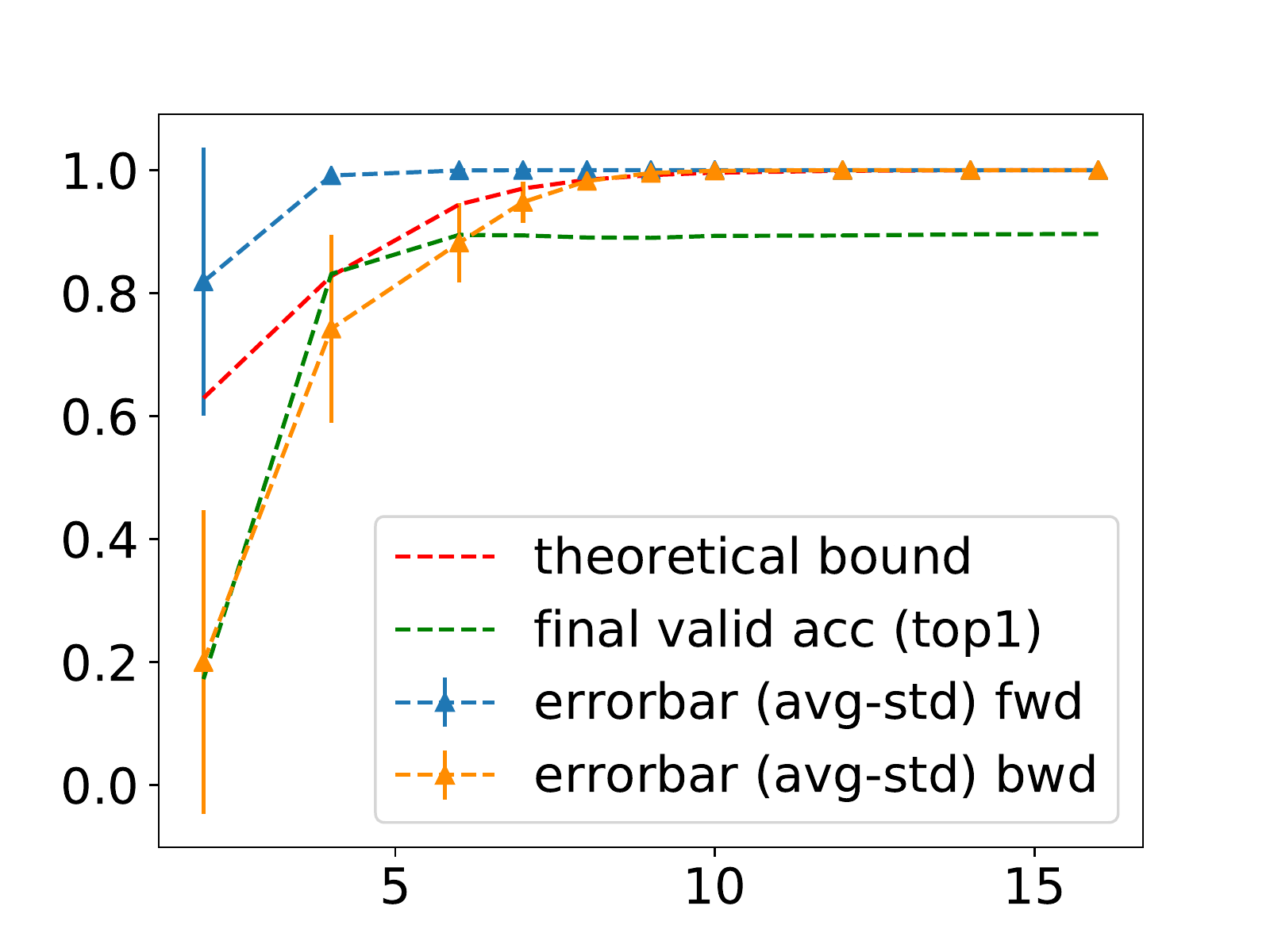}}
\end{subfigure}%
\hspace{5pt}
\begin{subfigure}{.5\textwidth}
\centerline{\includegraphics[width=\columnwidth]{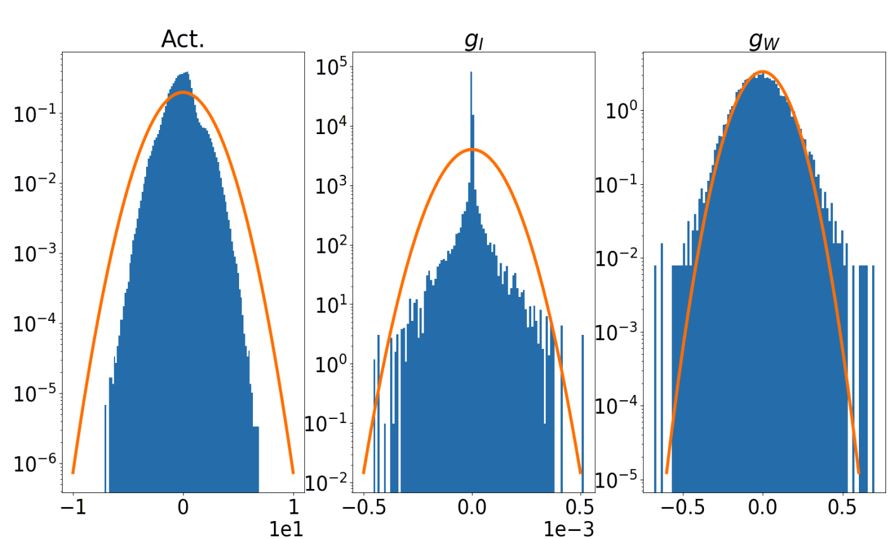}}

\end{subfigure}%
\caption{ Left: empirical and theoretical analysis of cosine similarity, $\cos(\theta)$, with respect the number of bits used for quantization. Right: Histograms of activations, layer gradients $g_l$ and weight gradients $g_W$. To emphasize that $g_l$ do not follow a Gaussian distribution, the histograms were plotted in a log-scale (ResNet-18, Cifar-10).}
\label{fig:emprical_analysis}
\end{figure}

To validate our assumption that the cosine similarity is a good measure for the quality of the quantization, we ran a set of experiments on Cifar-10 dataset, each with a different number of bits, and then plotted the average angle and the final accuracy. As can be seen in Figure \ref{fig:emprical_analysis} there is a high correlation between the two. Taking a closer look the following additional observations can be made:  (1) During quantization the direction of vectors is better preserved with the forward pass compared to the backward pass; (2) validation accuracy follows tightly the cosine of the angle in the backward pass, indicating gradient quantization as the primary bottleneck; (3)
as expected, the bound on $E(\cos(\theta))$ in Eq. \ref{eq:41} holds in the forward pass, but less so in the backward pass, where the Gaussian assumption tends to break. The histograms in Figure \ref{fig:emprical_analysis} further confirms that the layer gradients $g_l$ do not follow Gaussian distribution. These are the values that are bifurcated into low and high precision copies to reduce noise accumulation. 

% theoretical lower bound of $E(cos(\theta))$ established in Eq. \ref{eq:41} holds true for the forward propagation, but violated in the backward propagation, suggesting vectors involved in the backward pass are not Gaussian.  

%\begin{figure}[h]
%\scalebox{.5}
%
%\centerline{\includegraphics[width=1.2\columnwidth,heigh%t=0.35\columnwidth]{figures/angle_experiments_v2.pdf}}
%\caption{$cos(\theta)$ empirical and theoretical %analysis with respect the number of bits used for %quantization. Top row presents the cosine similarity for %vectors engaged in the forward pass while bottom row %present it for the backward pass. Left column shows the %cosine similarly error-bar ($\mu,\sigma$) with respect %to the number of bits. All experiments were conducted on %Cifar-10 dataset using ResNet-18 architecture. }
%\label{fig:emprical_analysis}
%\end{figure}

\begin{figure}[h] 
\vspace{-1em}

\begin{subfigure}{.5\textwidth}
\centerline{\includegraphics[width=\columnwidth]{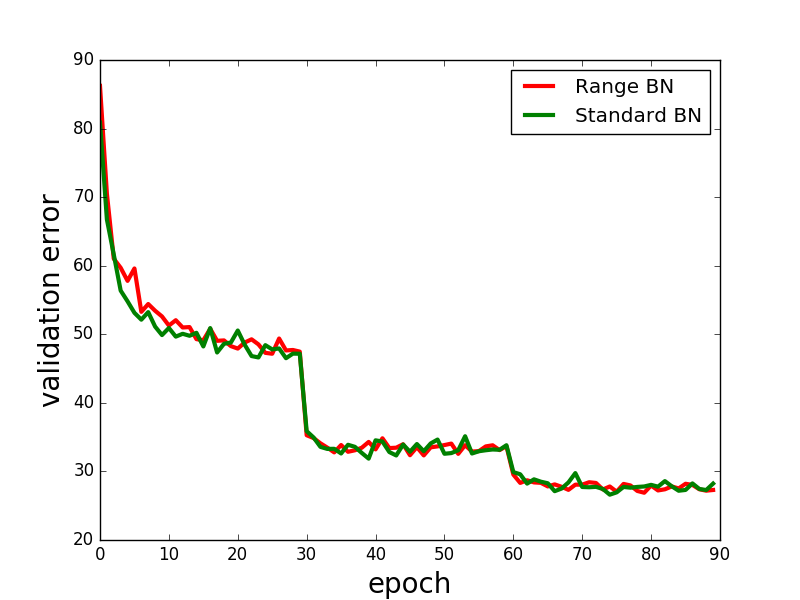}}
\end{subfigure}%
\begin{subfigure}{.5\textwidth}
\centerline{\includegraphics[width=\columnwidth]{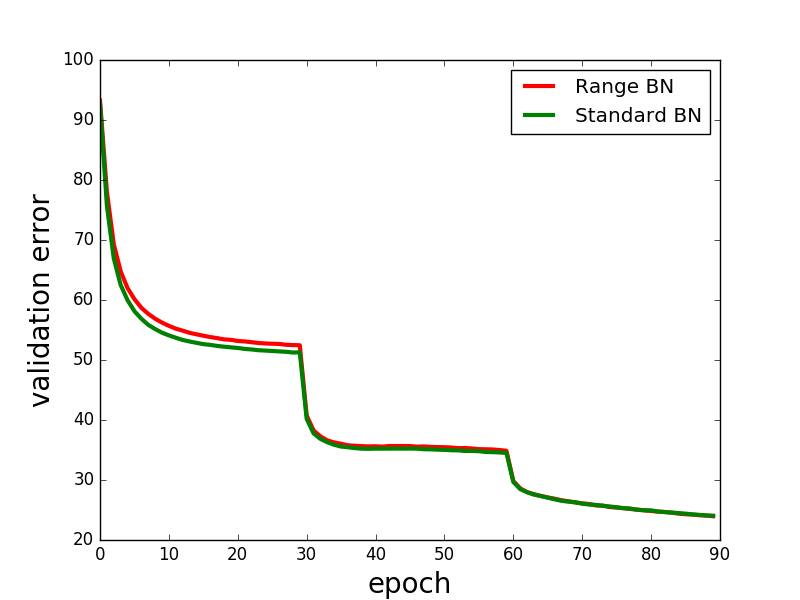}}
\end{subfigure}%
\caption{Equivalent accuracy with standard and range batch-norm (ResNet-50, ImageNet).}\label{range_Res50}
\vspace{-1em}
\end{figure}

\subsection{Experiment results on ImageNet dataset: Range Batch-Normalization}
We ran experiments with Res50 on ImageNet dataset showing the equivalence between the standard batch-norm and Range BN in terms of accuracy. The only difference between the experiments was the use of Range BN instead of the traditional batch-norm. Figure \ref{range_Res50} compares between the two. It shows equivalence when models are trained at high precision. We also ran simulations on other datasets and models. When examining the final results, both were equivalent i.e., 32.5\% vs 32.4\% for ResNet-18 on ImageNet and 10.5\% vs 10.7\% for ResNet-56 on Cifar10. 
To conclude, these simulations prove that we can replace standard batch-norm with Range BN while keeping accuracy unchanged. Replacing the sum of squares and square root operations in standard batch-norm by a few maximum and minimum operations has a major benefit in low-precision implementations.

\subsection{Experiment results on ImageNet dataset: Putting it all together}

\begin{figure}[h]
\vspace{-1em}

\begin{subfigure}{0.5\textwidth}
\centerline{\includegraphics[width=\columnwidth]{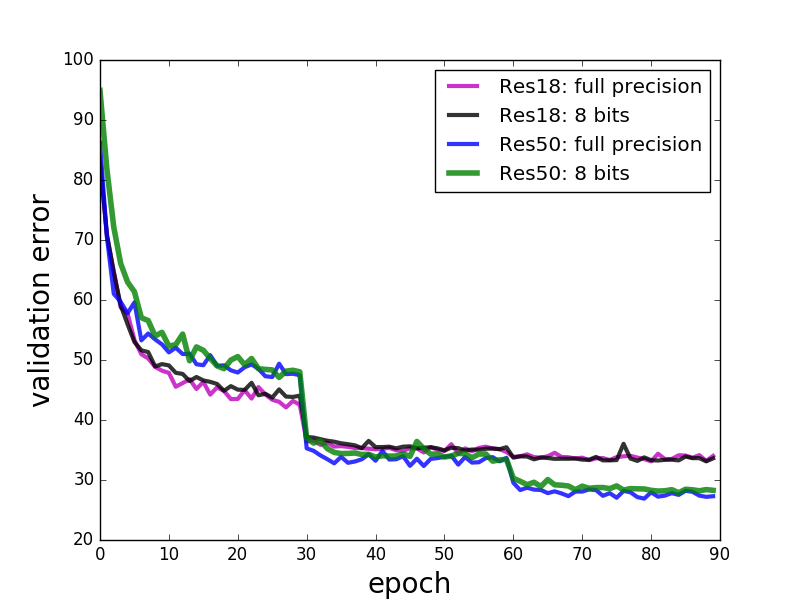}}
\end{subfigure}%
\begin{subfigure}{0.5\textwidth}
\centerline{\includegraphics[width=\columnwidth]{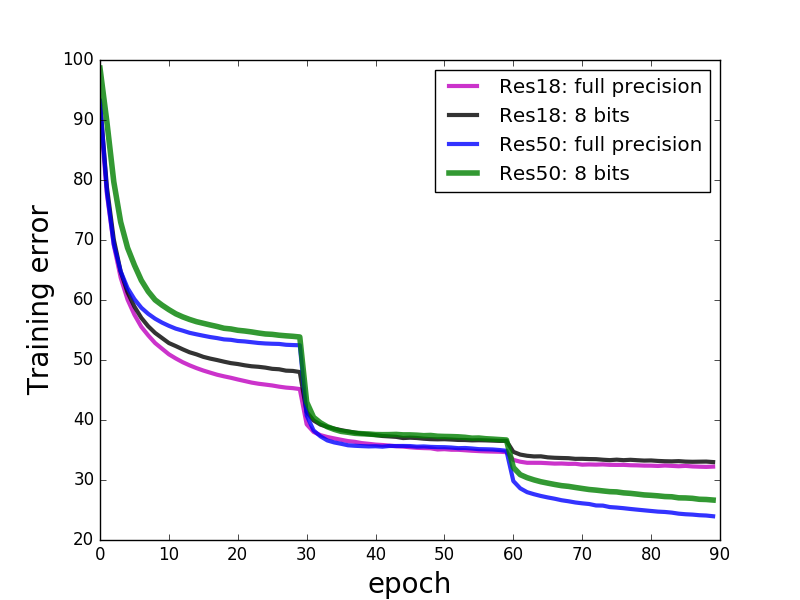}}
\end{subfigure}%
\caption{  Comparing a full precision run against  8-bit run with Qunatized Back-Propogarion and Range BN (ResNet-18 and ResNet-50 trained on ImageNet). }
\label{fig:Res18ImageNet}
\vspace{-1em}
\end{figure}

We conducted experiments using RangeBN together with Quantized Back-Propagation.  To validate this low precision scheme, we were quantizing the vast majority of operations to 8-bit. The only operations left at higher precising were the updates (float32) needed to accumulate small changes from stochastic gradient descent, and a copy of the layer gradients at 16 bits needed to compute $g_W$. Note that the float32 updates are done once per minibatch while the propagations are done for each example (e.g., for a minibatch of 256 examples the updates constitute less than ~0.4\% of the training effort). Figure \ref{fig:Res18ImageNet} presents the result of this experiment on ImageNet dataset using ResNet18 and ResNet50. We provide additional results using more aggressive quantizations in Appedix F.

\section{Discussion}

In this study, we investigate the internal representation of low precision neural networks and present guidelines for their quantization. Considering the preservation of direction during quantization, we analytically show that significant quantization is possible for vectors with a Gaussian distribution. On the forward pass the inputs to each layer are known to be distributed according to  a Gaussian distribution, but on the backward pass we observe that the layer gradients $g_l$ do not follow this distribution. Our experiments further assess that angle is not well preserved on the backward pass, and moreover final validation accuracy tightly follows that angle. Accordingly, we bifurcate the layer gradients $g_l$ and use it at 16-bits for the computation of the weight gradient $g_W$ while keeping the computation of next layer gradient $g_{l-1}$ at 8-bit. This enables the (slower) 16-bits computation of $g_W$ to be be done in parallel with $g_{l-1}$, without interrupting the propagation the layer gradients.

We further show that Range BN is comparable to the traditional batch norm in terms of accuracy and convergence rate. This makes it a viable alternative for low precision training. During the forward-propagation phase computation of the square and square root operations are avoided and replaced by $\max(\cdot)$ and $\min(\cdot)$ operations. During the back-propagation phase, the derivative of $\max(\cdot)$ or $\min(\cdot)$ is set to one where the coordinates for which the maximal or minimal values are attained, and is set to zero otherwise. 

Finally, we combine the two novelties into a single training scheme and demonstrate, for the first time, that 8-bit training on a large scale dataset does not harm accuracy. Our quantization approach has major performance benefits in terms of speed, memory, and energy. By replacing float32 with int8, multiplications become 16 times faster and at least 15 times more energy efficient \cite{hubara2016quantized}. This impact is attained for 2/3 of all the multiplications, namely the forward pass and the calculations of the layer gradients $g_l$. The weight gradients $g_W$ are computed as a product of 8-bit precision (layer input) with a 16-bit precision (unquantized version of $g_l$), resulting with a speedup of x8 for the rest of multiplications and at least x2 power savings. Although previous works considered an even lower precision quantization (up-to 1-bit),  we claim that 8-bit quantization may prove to be more of an interest. Furthermore, 8-bit matrix multiplication is available as an off-the-shelf operation in existing hardware and can be easily adopted and used with our methods.

\medskip

{
\small
\bibliography{nips_2018}
\bibliographystyle{nips-no-url}
}

\appendix
\part*{Appendix}

\section{Bound on the expected norm of $\epsilon$}\label{app:N-bit}

\remove{

\section{Angle preservation during quantization}

In this section we will give the full derivation of the proof provided in section 5. To estimate the angle between $W$ and $W+\epsilon$,  we first estimate the angle between $W$ and $\epsilon$. It is well known that if $\epsilon$ and $W$ are independent, then at high dimension the angle between $W$ and $\epsilon$ tends to $\dfrac{\pi}{2}$ \cite{biau2015high} i.e., we get a right angle triangle with  $W$ and $\epsilon$ as the legs, while $W+\epsilon$ is the hypotenuse as illustrated in Figure \ref{traingle}.

% In order to estimate the angle between $W$ and $W+\epsilon$,  we first estimate the angle between $W$ and $\epsilon$. It is well known that high dimension  if we draw two points at random from the unit ball, with high probability they (their vectors) will be nearly orthogonal in high dimensions i.e., with angle of $\dfrac{\pi}{2}\pm O(\dfrac{1}{\sqrt{N}})$ to each other \ref{biau2015high}. Since $\epsilon$ and $W$ are independent, and as the unit vectors $\dfrac{\epsilon}{||\epsilon||}$, $\dfrac{W}{||W||}$  are taken from the unit ball, we can conclude that the angle between $\dfrac{\epsilon}{||\epsilon||}$ and $\dfrac{W}{||W||}$ is nearly orthogonal (and so is the angle between $W$ and $\epsilon$). Therefore, by the definition of the vector addition operation, we get a right angle triangle with  $W$ and $\epsilon$ as the legs, while $W+\epsilon$ is the hypotenuse as illustrated in Figure \ref{traingle}. 

\begin{figure}[h] 
\begin{tikzpicture}[thick]
\coordinate (O) at (0,0);
\coordinate (A) at (12,0);
\coordinate (B) at (12,3);
\draw (O)--(A)--(B)--cycle;

\tkzLabelSegment[below](O,A){$W$}
\tkzLabelSegment(O,B){$W+\epsilon$}
\tkzLabelSegment[right](A,B){$\epsilon$}

\tkzMarkRightAngle[fill=orange,size=0.5,opacity=.4](O,A,B)% square angle here
\tkzLabelAngle[pos = 2](A,O,B){$\theta$}
\end{tikzpicture}
\caption{A geometric illustration of the angle $\theta$ at high dimensions}
\label{traingle}
\end{figure}

Therefore, the cosine of the angle $\theta$ can be approximated as follows:

\begin{equation}\label{eq:21}
\cos(\theta) =  \dfrac{||W||}{||W+\epsilon||}
\end{equation}

By the triangle inequality the following holds true:

\begin{equation}\label{eq:22}
\cos(\theta) =  \dfrac{||W||}{||W+\epsilon||}\geq \dfrac{||W||}{||W||+||\epsilon||}
\end{equation}

We now turn to estimate $||W||$ and $||\epsilon||$. First, since $W$ follows a Gaussian distribution with  a variance $\sigma^2$, the expected Euclidean norm  satisfies \cite{chandrasekaran2012convex}: 
\begin{equation}\label{eq:XX}
E(||W||)\cong \sqrt{N}\cdot \sigma
\end{equation}

}

By Jensen inequality the following holds true:
\begin{equation}\label{CorrectJensen}
E(||\epsilon||)= E\Bigg[\sqrt{\sum_{i=0}^{N-1} \epsilon_i^2} \Bigg] \leq \sqrt{E \Bigg[\sum_{i=0}^{N-1} \epsilon_i^2\Bigg]} = \sqrt{\sum_{i=0}^{N-1} E[\epsilon_i^2]}  
\end{equation}
Given a uniform random variable in $\epsilon_i \sim U[-\Delta/2,\Delta/2]$, we next derive the expected value of its square as follows:
\begin{equation}\label{Integral}
E[\epsilon_i^2]= \int_{-\Delta/2}^{\Delta/2} x^2\cdot\dfrac{1}{\Delta}\cdot dx = \dfrac{\Delta^2}{12}
\end{equation}
Note that $\Delta$ is a random variable. Hence, we substitute Eq \ref{Integral} into Eq \ref{CorrectJensen} using the following conditional expectation:
\begin{equation}\label{Uniform}
E(||\epsilon|| \text{ given } \Delta) \leq \sqrt{\frac{N}{12}}\cdot\Delta
\end{equation}
We can now establish $E(||\epsilon||)$ as follows:
\begin{equation}\label{eq:30}
E(||\epsilon||) \leq E (E(\,\,||\epsilon|| \,\,| \Delta)) \rvert \Delta))= \sqrt{\frac{N}{12}}\cdot E(\Delta)
\end{equation}

\remove{
%\begin{equation}\label{eq:30}
%E(||\Bar{\epsilon}||) = E_\Delta(||\epsilon|| \text{   given  } \Delta) = E_\Delta(\sqrt N\cdot\Delta \rvert \Delta)= \sqrt N\cdot E_\Delta(\Delta \rvert \Delta) = \sqrt N\cdot E(\Delta)
%\end{equation}

Finally, since $W$ follows a Gaussian distribution, and the expectation of the maximum of $N$ Gaussian random variables is upper bounded by  $\sigma\cdot\sqrt{2 \cdot \ln N}$ \citep{simon2007probability}, the following holds:

\begin{equation}\label{eq:31}
E(||\Bar{\epsilon}||) \leq \sqrt{\frac{N}{12}}\cdot E(\Delta) =  \sqrt{\frac{N}{12}}\cdot E\Bigg[\dfrac{\max(|W|)} {2^{M}} \Bigg] \leq  \dfrac{\sqrt N} {\sqrt{6}\cdot2^{M}}\cdot \sigma\cdot\sqrt{\ln N} 
\end{equation}

In high dimensions, the relative error made as considering $E(||X||_p)$ instead of the random value $||X||_p$ becomes asymptotically negligible \cite{biau2015high} (also known as the distance concentration phenomenon). Hence, we can replace the random variables in  Eq \ref{eq:22} by expectations and use Eq \ref{eq:XX} and Eq \ref{eq:31} as follows:

\begin{equation}\label{eq:99}
\cos(\theta) \geq \dfrac{||W||}{||W||+||\epsilon||}=\dfrac{E(||W||)}{E(||W||)+E(||\epsilon||)} \geq   \dfrac{2^{M}}{2^{M} + \sqrt{\ln N}/\sqrt{6}}
\end{equation}

% This together with the previous observations of  $E(||W||)\cong \sqrt{N}\cdot \sigma$ and Eq \ref{eq:31} establishes the following: 

% \begin{equation}\label{eq:24}
% \cos(\theta) \geq   \dfrac{2^{M}}{2^{M} + \sqrt{2/3 \cdot \log N}}
% %\geq \dfrac{2^{N-1}\cdot \sigma}{2^{N-1}\cdot \sigma+2\cdot \max(|W|)} 
% \end{equation}

% As expected equation \ref{eq:24} shows that large bit width $N$ increase the cosine similarity value (i.e., angle is smaller). Also, it is not very surprising to see that when  the bit width $N$ is fixed, larger values for $Max(|W|)$ reduces the cosine similarity (angle is larger).  Yet, it is somewhat surprising that larger $\sigma$ improves similarity. The later suggests that a  pre-processing of the weights $W$ before quantization (e.g., inference) should not take into account large outliers. For example, the removal of large outliers without significantly changing the standard deviation $\sigma$ should improve quantization. 

}

\section{Quantization methods}\label{app:GEMMLOWP}
 Following \cite{wu2016google} we choose to use the GMMLOWP quantization scheme as decribed in Google's open source library \citep{Benoit2017gemmlowp}. Given an input tensor $x$, clamping values $[v_{\min},v_{\max}]$,and number of bits $M$ we set the output to be:
\begin{align*}\label{eq:pareto mle2}
\mathrm{scale} &=   (v_{\max}-v_{\min})/2^M\\
\mathrm{zero}-\mathrm{point} &= \mathrm{round}(\min(\max(-v_{\min}/\mathrm{scale},0),2^M)) \\
\mathrm{output} &= \mathrm{round}(x/\mathrm{scale}+\mathrm{zero-point})\\
\end{align*}
The clamping values for the weights and activations were defined as the input's absolute maximum and minimum. Since the activations can have a high dynamic range which can be aggressively clamped as shown by \cite{wu2016google} we defined its clamping values to be the average of absolute maximum and minimum values of K chunks. This reduces the dynamic range Variance and allows smaller quantization steps. 

it is important to note that a good convergence was achieved only by using stochastic rounding \cite{gupta2015deep}. This behaviour is not surprisings as the gradients serves eventually for the weight update thus unbias quantization scheme is required to avoid quantization noise accumulation.

\section{Additional Experiments}
In this section we present our more aggressive quantization experiments of the Quantized Back-Propagation scheme. In the extreme case, QBP ternarizes the gradients and uses only 1-bit for the weights and activations. In this case, we refer to QBP networks as Ternarized Back-Propagation (TBP), in which  all forward MACs  operations  can  be  replaced  with XNOR and population count (\emph{i.e.,} counting the number of ones in the binary number) operations. To avoid significant degradation in test accuracy, we apply stochastic ternarization and increase the number of filter maps in a each convolution layer.

\subsection{CIFAR10}
A well studied dataset is the CIFAR10 image classification benchmark first introduced by \citet{krizhevsky2009learning}.  CIFAR10 consists of a training set of size 50K, and a test set of size 10K color images. Here, each images represents one of the following categories: airplanes, automobiles, birds, cats, deer, dogs, frogs, horses, ships and trucks.\\
We trained a VGG-like network similar to the one suggested by \citet{hubara2016binarized}  on the CIFAR10 dataset, with the same hyper-parameters used in the original work. We compared two variants: the original model BNN model and the BNN model trained with TBP (this paper), where ternarized gradients are used. \\
The results shown in Table (\ref{simple-table}) demonstrate that if we inflate the convolutional filters by 3 we can achieve similar results as the BNN and full precision models achieved. This is in accordance with previous finding \citep{mishra2017wrpn}, that found that widening the network can mitigate accuracy drop inflicted by low precision training. To make sure this is not a unique case for BNN we also applied TBP on ResNet with depth of 18. As can be seen from Table (\ref{simple-table}), as before, inflating the network improves performance, until it is only 1\% from the original performance, after inflating it by 5. 

\subsection{ImageNet}
Next, we applied TBP  to the more challenging ImageNet classification task introduced by \citet{deng2009imagenet}. It consists of a training set of size 1.2M samples and a test set of size 50K. Each instance is labeled with one of 1000 categories including objects, animals, scenes, and even some abstract shapes. We report two error rates for this dataset:  top-1 and top-5, as is typical done. Top-$k$ error rate represent the fraction of test images for which the correct label is not among the $k$ most probable labels predicted by the model.  

We run several experiments on AlexNet inflated by 3. Similarly to previous work and to ease the comparison we kept first and last layer in full precision. With binarized weights, 4-bit activations and gradients, TBP converged to 53.3\% top-1 accuracy and 75.84\% top-5 accuracy. By using only 2bit activation TBP reached 49.6\% top-1 accuracy and 73.1\% top-5 accuracy. We are currently working on more advanced typologies such as ResNet-50 model \cite{he2016deep}. Results are summarized in Table (\ref{tab:ImageNet}).

\begin{table}[ht]
\caption{Classification top-1 validation error rates of TBP BNNs trained on ImageNet with AlexNet topology. I, A, W, G stands for Inflation, Activation bits, Weights bits and Gradient bits respectively}
\label{tab:ImageNet}
\vskip 0.15in
\begin{center}
\begin{small}
\begin{sc}

\begin{tabular}{lccccr}
\toprule
Model & I & A & W & G &Error \\
\midrule
TBP (ours)    & 3 & 2& 1& 4& 50.43\%  \\
TBP (ours)   & 3 & 4& 1& 4& 46.7\%  \\
DoReFa (\citeauthor{zhou2016dorefa}) & 1 & 2& 1& 6& 53.9\%  \\
DoReFa (\citeauthor{zhou2016dorefa}) & 1 & 4& 1& 6& 51.8\%  \\
WRPN (\citeauthor{mishra2017wrpn}) & 2 & 1& 1& 32& 48.3\%  \\
WRPN (\citeauthor{mishra2017wrpn}) & 2 & 2& 2& 32& 55.8\%  \\
\hline\hline
\multicolumn{5}{c}{Single Precision -No quantiztion}\\
\hline\hline
AlexNet (\citeauthor{krizhevsky2014one}) & 1 & 32& 32& 32& 43.5\%  \tabularnewline
\bottomrule
\end{tabular}

\end{sc}
\end{small}
\end{center}
\end{table}

\begin{table}[ht]
\caption{Classification test error rates of  TBP BNNs trained on CIFAR10}
\label{simple-table}
\vskip 0.15in
\begin{center}
\begin{small}
\begin{sc}
\begin{tabular}{lcccr}
\toprule

Model&  Error Rate\\
\midrule
\multicolumn{5}{c}{Binarized activations,weights and tern gradients}
\tabularnewline
\hline 
\hline 
TBP  BNN, inflated by 3             & 9.53\%  & \tabularnewline
TBP  ResNet-18, inflated by 5       & 10.8\%  & \tabularnewline
TBP  ResNet-18, inflated by 3       & 14.21\%  & \tabularnewline
TBP  ResNet-18 \citep{he2016deep} & 18.5\%  & \tabularnewline
\hline 
\hline
\multicolumn{5}{c}{Binarized activations,weights}\tabularnewline
\hline 
\hline 
BNN \cite{hubara2016binarized}     & 10.15\%  & \tabularnewline
Binarized ResNet-18, inflated by 5   & 10.7\%  & \tabularnewline
\multicolumn{5}{c}{No binarization (standard results)}\tabularnewline
\hline 
\hline 
BNN \cite{hubara2016binarized}     & 10.94\% & \tabularnewline
ResNet-18                          & 9.64\% & \tabularnewline
\hline 
\bottomrule
\end{tabular}
\end{sc}
\end{small}
\end{center}
\vskip -0.1in
\end{table}

\subsection{Additional experiments}
To shed light on why TBP works, and on what is yet to be solved, we conducted additional set of experiments on the CIFAR-10 dataset with both ResNet-18 and the BNN like typologies.
The Back Propagation algorithm includes three phases. The forward phase in which we calculate the activations values for each layer. The Backward-Propagation (BP) phase in which the gradients from layer $\ell$ pass to layer $\ell-1$ and the update phase in which the model parameters (\emph{i.e.,} weights) are updated. Both stages require MAC operations as detailed in section \ref{sec:QBP}. In this paper we focused on the BP stage. As oppose to the update stage that can be done in parallel, BP is a sequential stage. However, if the update phase uses full precision MAC operation the hardware need to support it. Moreover compressing the update gradients reduce the communication bandwidth in distributed systems. Thus, quantizing the weights gradients for the updates phase can also reduce power consummation and accelerate the training.  
% \dnote{you should clarify (here and before) what you mean by the precision of the updates and the gradients, why is it important, why did we do so far (only gradients), why is this reasonable, but also why should we try to improve and also quantize the updates. Without this explanation, this section is very unclear, and it is not clear why do we such low results in table 2, where until now the results were much better.}
\paragraph{Ternarizing both stages.}
%\dnote{what tensor? this is the first time you mention tensors in the paper, and so it is not clear what are you talking about.} 
Ternarizing both stages results with completely MAC free training. However, our results show that without enabling at least 3bit precision for the update stage the model reaches only approximately 80\% accuracy. This indicates that the ternarization noise is too high, and thus distorts the update gradients direction. If we stop the gradients ternarization once the accuracy ceases to increase, the convergence continues and the accuracy increases to the same accuracy as TBP. Thus, ternarizing the update stage can be used to accelerate TBP training of BNN networks by first training it with ternarized weights gradients and  then, for the last couple of epochs, continue training with full precision weights gradients.
%\dnote{sentence not clear.}. 
%\dnote{sentence not clear. Also, what is "update angle"?}.
%Itay: Is that clearer, feel free to change or delete this sentence
\begin{algorithm}[h]
\begin{algorithmic}
    \REQUIRE gradients from previous layer $g_{s_k}$, binarized activation $a_{k-1}^{b}$  and number of samples $S$.
    \FOR{$k=1$ to $S$} 
        \STATE $g_{s_k}^{b} \leftarrow {\rm StcTern}(g_{s_k})$
        \STATE $g_{W_k^b} \mathrel{+}= {g_{s_k}^b} a_{k-1}^{b}$
    \ENDFOR \\
    \STATE \bf{Return} $g_{W_k^b} \mathrel{/}= S$
\caption{Multiple stochastic ternarization sampling algorithm.}\label{algoritm:multi}   
\end{algorithmic}
\end{algorithm}

\begin{table}[!]
\begin{center}
\begin{tabular}{lcccr}

\hline 
Data set & Error\\
\hline 
\hline 
TBP  BNN, 1 sample             & 20.1\%  & \tabularnewline
TBP  BNN, 5 samples            & 13.5\%  & \tabularnewline
TBP  BNN, 10 samples           & 13\%    & \tabularnewline
TBP  BNN, 20 samples           & 12\%  & \tabularnewline
TBP  BNN, $R>0.7$              & 12.5\%  & \tabularnewline
ResNet-18 10 samples    & 14\%  & \tabularnewline
ResNet-18 20 samples    & 12.7\%  & \tabularnewline
\hline 
\bottomrule
\end{tabular}
\vspace{10pt}
 \caption{Classification test error rates of  TBP BNNs trained on CIFAR10 with Multiple Stochastic ternarization sampling for the update phase. ResNet-18 and BNN models were inflated by 5 and 3 respectively.\label{tab:multi}}
\end{center}
\end{table}
\paragraph{Multiple stochastic ternarization sampling.}
To alleviate the need for float MAC operation in the update phase we suggest to use gated XNOR operation multiple times, each time with different stochastic sample of the ternarized tensor and average the results. The algorithm is detailed in Algorithm (\ref{algoritm:multi}) and results are given in Table (\ref{tab:multi}). As expected the accuracy improves with the amount of sampling. To find the number of samples needed for each layer we adopted  a similar  geometrical approach as suggested by \citet{anderson2017high} and measured the correlation coefficient ($R$) between the update gradients received with and without ternarization. Our experiments indicate that  more samples are required for the first two convolution layer (12 samples) while the rest of the layers need approximately 6 samples. Using this configuration keeps the correlation coefficient above 0.7 and results with 87.5\% accuracy.

%\subsection{Selecting the optimal quantization step %$\Delta$}
%
%In this subsection we propose a quantization method  to %compute the optimal quantization step $\Delta$ for $k$ %bit representation. Given a tensor of weights $W$ in the %following we show that $\Delta_{opt} = %\dfrac{2^k\cdot\max (W)}{N+2^{2k}}$. 
%
%The mean-square-error (MSE) between the quantified and %full precision tensor can be decomposed into variance  %and bias squared as follows [ref]:
%
%
%\begin{equation}\label{eq:50}
%\text{MSE }(W - Q(W)) = \dfrac{1}{N}} ||W - Q(W)||^2 = %\text{mean}(W - Q(W))^2 + \text{var}(W - Q(W))
%\end{equation}
%
%Our goal would be to minimize the expected value of MSE. %We assume 
%
%We assume for the sake of simplicty that we have a %single outlier 

\end{document}


\maketitle

% \begin{abstract}

% \bigskip
% \end{abstract}
\setcounter{page}{1}
% \thispagestyle{empty}
% \end{titlepage}
% \pagebreak \newpage

\appendix

\dnote{I would re-sturcute and put all the theoretical sections A,B,C,D as one section with subsections. Also, give some intro sentence to explain what is this sectin about.}
\section{Problem Statement} \label{sec:Problem}

We are given a vector of weights $W=(w_0,w_1,...,w_{n-1})$, where weights follow a Gaussian distribution  $W \sim N(0,\sigma)$. We would like to measure the cosine similarity (i.e., cosine of the angle) between $W$ and $Q(W)$, where $Q(\cdot)$ is a quantization function. More formally, we are interested in estimating the following geometric measure:

\begin{equation}\label{eq:1}
\cos(\theta)=\dfrac{W\cdot Q(W)}{||W||_2\cdot ||Q(W)||_2}
\end{equation}

In the following we consider the case where  $Q(\cdot)$ is binary, ternary and finally uniform n-bit quantization.

\section{Binary Quantization} 

Here we assume that $Q(\cdot)$ is a a mapping $Q\colon\mathbb{R}\to\{-1 , 1\}$ whose behavior in each coordinate is as per the sign function i.e., 

\begin{equation}
(w_0,w_1,...,w_{n-1})\to (\text{sgn}(w_0),\text{sgn}(w_1),...,\text{sgn}(w_{n-1}))
\end{equation}

Considering the numerator of equation \ref{eq:1}, as the product of each coordinate by its sign results with a positive value, the dot product $W\cdot Q(W)$ can be expressed by the following $L^1$ norm representation:

\begin{equation}\label{eq:3}
W\cdot Q(W) = \sum_{i=0}^{n-1} |w_i| = ||W||_1
\end{equation}

We turn to consider the denominator of equation \ref{eq:1}. First, it is easy to see that since $Q(W)$ is a $n$ dimensional vector with coordinates either at  -1 or 1, the following holds true:

\begin{equation}
Q(W) = \sqrt{\sum_{i=0}^{n-1} w_i^2} =  \sqrt{\sum_{i=0}^{n-1} 1} = \sqrt{n}
\end{equation}

Therefore for the binary case we can substitude equation \ref{eq:1} as follows:

\begin{equation}\label{eq:2}
\cos(\theta)=\dfrac{||W||_1}{||W||_2\cdot \sqrt{n}}
\end{equation}

We now employ the assumption that $W \sim N(0,\sigma)$, with the purpose of establishing the expected cosine of the angle between $W$ and $Q(W)$ using the expression given at  \ref{eq:2}. Note that since the function $1/x$ is convex for $x\geq 0$, we can apply Jensen's inequality for the random variable $||W||_2$ to achieve a lower bound on the expected cosine angle of Equation \ref{eq:2} as follows.

\begin{equation}\label{eq:6}
E\Bigg( \dfrac{||W||_1}{||W||_2\cdot \sqrt{n}}\Bigg) \geq E\Bigg( \dfrac{||W||_1}{E(||W||_2)\cdot \sqrt{n}}\Bigg)=\dfrac{E(||W||_1)}{E(||W||_2)\cdot \sqrt{n}}
\end{equation}

Hence, to conclude this analysis we still need to  establish the expected value of the random variables $||W||_1$ and $||W||_2$. Considering first  $E(||W||_1)$, by the linearity of expectation the following must hold.

\begin{equation}
E(||W||_1)=E\Big(\sum_{i=0}^{n-1} |w_i| \Big)=\sum_{i=0}^{n-1} E(|w_i|)
\end{equation}

Next, as each $w_i$ belongs to a normal distribution with a mean $\mu=0$ and variance $\sigma$, the absolute random variable $|w_i|$ has a  folded normal distribution with a mean ${\mu}_2$ as follows :
\begin{equation}
{\mu}_2 = \sigma\sqrt{\dfrac{2}{\pi}} \exp\Bigg(\dfrac{-\mu}{2\sigma^2} \Bigg)
\end{equation}

Therefore, as $\mu=0$ we can obtain that ${\mu}_2 =\sigma \sqrt{\dfrac{2}{\pi}}$, which can be summarized as follows:

\begin{equation}\label{eq:9}
E(||W||_1)=\sum_{i=0}^{n-1} E(|w_i|)=\sum_{i=0}^{n-1} E(|w_i|)=\sigma \sqrt{\dfrac{2}{\pi}}=n\cdot\sigma \sqrt{\dfrac{2}{\pi}}
\end{equation}

We turn to consider $E(||W||_2)$. As stated in  [The Convex Geometry of Linear Inverse Problems] we know that the average $L_2$ norm of a $N$-dimensional vector given by a normal distribution $N(0,\sigma)$ satisfies the following: 

\begin{equation}\label{eq:10}
E(||W||_2)\leq \sqrt{N}\cdot \sigma
\end{equation}

Finally, by substituting equations \ref{eq:9} and \ref{eq:10} into equation \ref{eq:6} we can conclude this analysis as follows.

\begin{equation}
E\Bigg( \dfrac{||W||_1}{||W||_2\cdot \sqrt{n}}\Bigg) \geq \dfrac{E(||W||_1)}{E(||W||_2)\cdot \sqrt{n}}\geq \sqrt{\dfrac{2}{\pi}} 
\end{equation}

This establishes the fact that the angle $\theta$ is (on expectation) upper bounded by a $arccos \sqrt{\dfrac{2}{\pi}} \sim 37$ degrees. 

\section{Ternary Quantization} 

Given a treshold $t$, in this subsection the binary quantization function is replaced by the following ternary quantization function $Q(x)=1$ if $x>t$, 0 if $|x|<t$, and -1 if $x<-t$. In the following we provide a close form solution for the expected value of the cosine angle for any threshold t. Considering equation \ref{eq:1}, the cosine measure is convex with respect to the random variables $||W||_2$ and $||Q(W)||_2$. Therefore the following derivation follows directly by applying Jensen's inequality as follows:
\begin{equation}\label{eq:12}
E\Bigg( \dfrac{W\cdot Q(W)}{||W||\cdot ||Q(W)||}\Bigg) \geq E\Bigg( \dfrac{W\cdot Q(W)}{E(||W||)\cdot E(||Q(W)||)}\Bigg)=\dfrac{E(W\cdot Q(W))}{E(||W||)\cdot E(||Q(W)||)}
\end{equation}
We next turn to find the expectation of each term in equation \ref{eq:12}, and  begin with the numerator. Following similar arguments to the one given for equation \ref{eq:2}, it is easy to see that the following holds true.

\begin{equation}\label{eq:13}
E(W\cdot Q(W))= E\Big(\sum_{i}^{} |w_i| \Big)=\sum_{i, |w_i|>t}^{} E(|w_i| )
\end{equation}

Since $w_i$ follows the Gaussian distribution $N(0,\sigma)$, the proportion of values that satisfy $ |w_i|<t$ are given by $\Phi(t)-\Phi(-t)$, where $\Phi(\cdot)$ is the cumulative distribution function of the standard normal distribution $\Phi(x)=\dfrac{1}{2}[1+\text{erf}(x/\sqrt2)]$. By the symmetry of  $\Phi(\cdot)$ around zero, we can conclude that $\Phi(t)-\Phi(-t)=2\Phi(t)-1$. Hence, the probability of $|w_i| \geq t$ equals to $1-[2\Phi(t)-1]=2-2\Phi(t)$. Accordingly, from equation \ref{eq:13}, the following can be  established:

\begin{equation}\label{eq:14}
E(W\cdot Q(W))= \sum_{i, |w_i|>t}^{} E(|w_i| )= (2-2\Phi(t))\cdot \sum_{i}^{} E(|w_i|, |w_i|>t)=(2-2\Phi(t))\cdot \sum_{i}^{} E(w_i, w_i>t)
\end{equation}

The last derivation is a property that follows from the symmetry of the Gaussian distribution around zero i.e., $E(w_i, w_i>t)=-E(w_i, w_i<-t)$. Finally, the expected value of random variable $w_i$ with the condition that $w_i>t$ can be derived using the truncated Gaussian distribution as follows:

%Gaussian distribution: $\phi(x) = \frac{1}{{\sigma \sqrt {2\pi } }}e^{{{ - \left( {x } \right)^2 } \mathord{\left/ {\vphantom {{ - \left( {x } \right)^2 } {2\sigma ^2 }}} \right. \kern-\nulldelimiterspace} {2\sigma ^2 }}}$

\begin{equation}\label{eq:15}
E(w_i, w_i>t) =  E(w_i)+\dfrac{\phi(t)}{1-\Phi(t)}\cdot\sigma=\dfrac{\phi(t)}{1-\Phi(t)}\cdot\sigma
\end{equation}

where $\phi(t)$ is the probability density function   $\phi(x) = \frac{1}{{\sigma \sqrt {2\pi } }}e^{{{ - \left( {x } \right)^2 } \mathord{\left/ {\vphantom {{ - \left( {x } \right)^2 } {2\sigma ^2 }}} \right. \kern-\nulldelimiterspace} {2\sigma ^2 }}}$. Therefore, by substituting equation \ref{eq:15} into equation \ref{eq:14} we obtain the following:

\begin{equation}\label{eq:17}
E(W\cdot Q(W))= (2-2\Phi(t))\cdot \sum_{i}^{} E(w_i, w_i>t)= \dfrac{\phi(t)\cdot(2-2\Phi(t))}{1-\Phi(t)}\cdot n \cdot\sigma = 2n\sigma\phi(t)
\end{equation}

We now turn to estimate $E(||Q(W)||)$. Since the probability of each coordinate having non-zero value is $2-2\Phi(t)$, and as non-zero values are restricted to either 1 or -1, it is easy to show that the following holds:

\begin{equation}\label{eq:16}
E(||Q(W)||) =  \sqrt{n(2-2\Phi(t))}
\end{equation}

Finally, employing again the expectation of $E(||W||)$ from Eq. \ref{eq:10}, we can substitute into equation \ref{eq:12} the expectations established in Eq. \ref{eq:16} and  Eq.\ref{eq:17}:

\begin{equation}\label{eq:18}
E\Bigg( \dfrac{W\cdot Q(W)}{||W||\cdot ||Q(W)||}\Bigg) \geq \dfrac{E(W\cdot Q(W))}{E(||W||)\cdot E(||Q(W)||)}\geq \dfrac{2\phi(t)}{\sqrt{2-2\Phi(t)}}
\end{equation}

As a sanity check, it is easy to validate that the expected angle in the case when $t=0$ (for which the ternary and binary quantization schemes are equivalent) is indeed $arccos \sqrt{\dfrac{2}{\pi}} \sim 37$ degrees. Specifically, as $\phi(0)  = \dfrac{1}{\sqrt{2\pi}}$ and $\Phi(0)  = 0.5$ [ref],  the following must hold true.

\begin{equation}
E\Bigg( \dfrac{W\cdot Q(W)}{||W||\cdot ||Q(W)||}\Bigg) \geq \dfrac{2\phi(0)}{\sqrt{2-2\Phi(0)}} = \sqrt{\dfrac{2}{\pi}}
\end{equation}

\begin{figure}[h]
\vskip 0.2in
\begin{center}
\centerline{\includegraphics[width=\columnwidth]{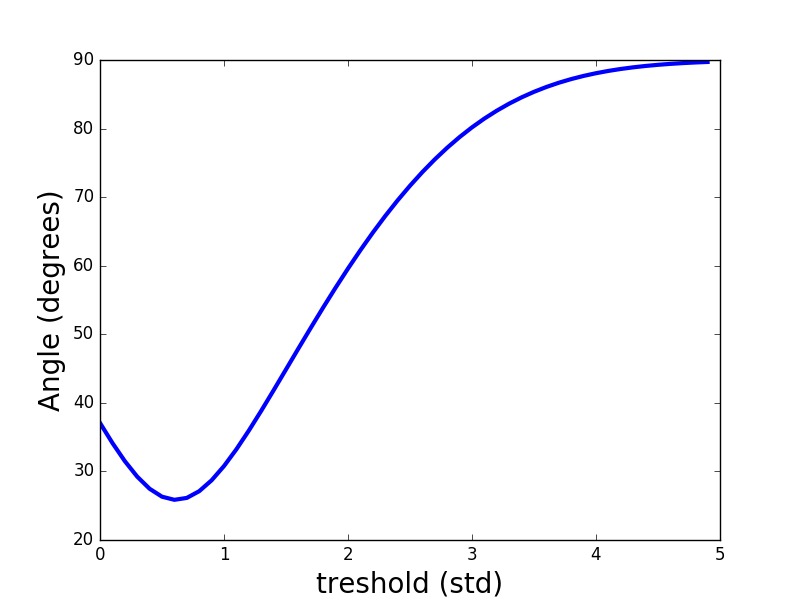}}
\caption{Angle vs. Threshold as predicted by equation \ref{eq:18}. Note that for t=0 we collapse to the binary settings where the angle is 37 degrees while for larger  thresholds too many values are zeroed hampering the correlation between the high precision vector $W$ and its ternarized version $Q(W)$. Minimum angle  is 25.8 degrees for a threshold of 0.6 standard deviations.}
\label{tershVsAngle}
\end{center}
\vskip -0.2in
\end{figure}

Figure \ref{tershVsAngle} presents the angle (degrees) versus the threshold t (in standard deviation units) as predicted by Eq. \ref{eq:18}

\section{$N$-bit quantization}

In this section we consider the case of $M$ bit quantization. To that end, we define a fixed quantization step between adjacent quantified levels  as follows:

\begin{equation}
\Delta= \dfrac{\max(|W|)}{2^{M}}
\end{equation}

 Given a vector of quantization error $\Bar{\epsilon}$ taken from a uniform distribution (i.e., for each index $i$ in  $\Bar{\epsilon}= (\epsilon_0,\epsilon_1,...,\epsilon_{n-1})$, the coordinate $\epsilon_i$ follows the uniform distribution $\mathcal{U}[-\Delta/2,\Delta/2]$), the quantized weights are modeled as a superposition of $W$ and $\Bar{\epsilon}$ i.e.,   $Q(W) = W + \Bar{\epsilon}$.

In order to estimate the angle between $W$ and $W+\epsilon$,  we first estimate the angle between $W$ and $\epsilon$. It is well known that  if we draw two points at random from the unit ball, with high probability they (their vectors) will be nearly orthogonal in high dimensions i.e., with angle of $\dfrac{\pi}{2}\pm O(\dfrac{1}{\sqrt{N}})$ to each other. Since $\epsilon$ and $W$ are independent, and as the unit vectors $\dfrac{\epsilon}{||\epsilon||}$, $\dfrac{W}{||W||}$  are taken from the unit ball, we can conclude that the angle between $\dfrac{\epsilon}{||\epsilon||}$ and $\dfrac{W}{||W||}$ is nearly orthogonal (and so is the angle between $W$ and $\epsilon$). Therefore, by the definition of the vector addition operation, we get a right angle triangle with  $W$ and $\epsilon$ as the legs, while $W+\epsilon$ is the hypotenuse as illustrated in Figure \ref{traingle}.

\begin{figure}[h] 
\begin{tikzpicture}[thick]
\coordinate (O) at (0,0);
\coordinate (A) at (12,0);
\coordinate (B) at (12,3);
\draw (O)--(A)--(B)--cycle;

\tkzLabelSegment[below](O,A){$W$}
\tkzLabelSegment(O,B){$W+\epsilon$}
\tkzLabelSegment[right](A,B){$\epsilon$}

\tkzMarkRightAngle[fill=orange,size=0.5,opacity=.4](O,A,B)% square angle here
\tkzLabelAngle[pos = 2](A,O,B){$\theta$}
\end{tikzpicture}
\caption{geometric illustration of the angle $\theta$ at high dimensions}
\label{traingle}
\end{figure}

Therefore, at high dimension the cosine of the angle $\theta$ can be approximated as follows:

\begin{equation}\label{eq:21}
\cos(\theta) =  \dfrac{||W||}{||W+\epsilon||}
\end{equation}

By the triangle inequality the following holds true:

\begin{equation}\label{eq:22}
\cos(\theta) =  \dfrac{||W||}{||W+\epsilon||}\geq \dfrac{||W||}{||W||+||\epsilon||}
\end{equation}

Since $\Bar{\epsilon}$ follows a uniform distribution in $\mathcal{U}[-\Delta/2,\Delta/2]$), the conditional expectation of  $||\Bar{\epsilon}||$ given $\Delta$ is $O(\sqrt N\cdot\Delta)$. We can therefore conclude the following (I am abusing the notation of big O for the sake of simplicity - that should be cleaned later)

\begin{equation}\label{eq:30}
E(||\Bar{\epsilon}||) = E_\Delta(||\epsilon|| \text{   given  } \Delta) = E_\Delta(\sqrt N\cdot\Delta \rvert \Delta)= \sqrt N\cdot E_\Delta(\Delta \rvert \Delta) = \sqrt N\cdot E(\Delta)
\end{equation}

Since $W$ follows a Gaussian distribution, and the expectation of the maximum of $N$ Gaussian random variables is upper bounded by  $\sigma\cdot\sqrt{2\cdot logN}$ [Simon 2007], the following holds:

\begin{equation}\label{eq:31}
E(||\Bar{\epsilon}||) = \sqrt N\cdot E(\Delta) =  \sqrt N\cdot E\Bigg[\dfrac{\max(|W|)} {2^{M}} \Bigg] \leq  \dfrac{\sqrt N} {2^{M}}\cdot \sigma\cdot\sqrt{2\cdot logN} 
\end{equation}

Hence, following Jensen inequality we can calculate the expected cosine similarity as follows:  

\begin{equation}\label{eq:23}
E \Bigg( \dfrac{||W||}{||W+\epsilon||} \Bigg)\geq E \Bigg( \dfrac{||W||}{||W||+||\epsilon||}\Bigg)\geq E\Bigg( \dfrac{||W||}{E(||W||)+E(||\epsilon||)}\Bigg)=  \dfrac{E(||W||)}{E(||W||)+E(||\epsilon||)}
\end{equation}

This together with the previous observations of  $E(||W||)\cong \sqrt{N}\cdot \sigma$ and Eq \ref{eq:31} establishes the following:

\begin{equation}\label{eq:24}
E \Bigg( \dfrac{||W||}{||W+\epsilon||} \Bigg)\geq   \dfrac{2^{M}}{2^{M} + \sqrt{2 \cdot log N}}
%\geq \dfrac{2^{N-1}\cdot \sigma}{2^{N-1}\cdot \sigma+2\cdot \max(|W|)} 
\end{equation}

% As expected equation \ref{eq:24} shows that large bit width $N$ increase the cosine similarity value (i.e., angle is smaller). Also, it is not very surprising to see that when  the bit width $N$ is fixed, larger values for $Max(|W|)$ reduces the cosine similarity (angle is larger).  Yet, it is somewhat surprising that larger $\sigma$ improves similarity. The later suggests that a  pre-processing of the weights $W$ before quantization (e.g., inference) should not take into account large outliers. For example, the removal of large outliers without significantly changing the standard deviation $\sigma$ should improve quantization. 

\section{Quantization methods}
 Following \cite{wu2016google} we choose to use the GMMLOWP quantization scheme as decribed in Google's open source library \citep{Benoit2017gemmlowp}. Given an input tensor $x$, clamping values $[v_{min},v_{max}]$,and number of bits $M$ we set the output to be:

\begin{align*}\label{eq:pareto mle2}
scale &=   (v_{max}-v_{min})/2^M\\
zero-point &= round(min(max(-v_{min}/scale,0),2^M)) \\
output &= round(x/scale+zero-point)\\
\end{align*}

The clamping values for the weights and activations were defined as the input's absolute maximum and minimum. Since the activations can have a high dynamic range which can be aggressively clamped as shown by \cite{wu2016google} we defined its clamping values to be the average of absolute maximum and minimum values of K chunks. This reduces the dynamic range variance and allows smaller quantization steps. 

it is important to note that a good convergence was achieved only by using stochastic rounding \cite{gupta2015deep}. This behaviour is not surprisings as the gradients serves eventually for the weight update thus unbias quantization scheme is required to avoid quantization noise accumulation.

\section{Additional Experiments}
In this section we present our more aggressive quantization experiments of the Quantized Back-Propagation scheme. In the extreme case, QBP ternarizes the gradients and uses only 1-bit for the weights and activations. In this case, we refer to QBP networks as Ternarized Back-Propagation (TBP), in which  all forward MACs  operations  can  be  replaced  with XNOR and population count (\emph{i.e.,} counting the number of ones in the binary number) operations. To avoid significant degradation in test accuracy, we apply stochastic ternarization and increase the number of filter maps in a each convolution layer.

\subsection{CIFAR10}
A well studied dataset is the CIFAR10 image classification benchmark first introduced by \citet{krizhevsky2009learning}.  CIFAR10 consists of a training set of size 50K, and a test set of size 10K color images. Here, each images represents one of the following categories: airplanes, automobiles, birds, cats, deer, dogs, frogs, horses, ships and trucks.\\
We trained a VGG-like network similar to the one suggested by \citet{hubara2016binarized}  on the CIFAR10 dataset, with the same hyper-parameters used in the original work. We compared two variants: the original model BNN model and the BNN model trained with TBP (this paper), where ternarized gradients are used. \\
The results shown in Table (\ref{simple-table}) demonstrate that if we inflate the convolutional filters by 3 we can achieve similar results as the BNN and full precision models achieved. This is in accordance with previous finding \citep{mishra2017wrpn}, that found that widening the network can mitigate accuracy drop inflicted by low precision training. To make sure this is not a unique case for BNN we also applied TBP on ResNet with depth of 18. As can be seen from Table (\ref{simple-table}), as before, inflating the network improves performance, until it is only 1\% from the original performance, after inflating it by 5. 

\subsection{ImageNet}
Next, we applied TBP  to the more challenging ImageNet classification task introduced by \citet{deng2009imagenet}. It consists of a training set of size 1.2M samples and a test set of size 50K. Each instance is labeled with one of 1000 categories including objects, animals, scenes, and even some abstract shapes. We report two error rates for this dataset:  top-1 and top-5, as is typical done. Top-$k$ error rate represent the fraction of test images for which the correct label is not among the $k$ most probable labels predicted by the model.  

We run several experiments on AlexNet inflated by 3. Similarly to previous work and to ease the comparison we kept first and last layer in full precision. With binarized weights, 4-bit activations and gradients, TBP converged to 53.3\% top-1 accuracy and 75.84\% top-5 accuracy. By using only 2bit activation TBP reached 49.6\% top-1 accuracy and 73.1\% top-5 accuracy. We are currently working on more advanced typologies such as ResNet-50 model \cite{he2016deep}. Results are summarized in Table (\ref{tab:ImageNet}).

\begin{table}[ht]
\caption{Classification top-1 validation error rates of TBP BNNs trained on ImageNet with AlexNet topology. I, A, W, G stands for Inflation, Activation bits, Weights bits and Gradient bits respectively}
\label{tab:ImageNet}
\vskip 0.15in
\begin{center}
\begin{small}
\begin{sc}

\begin{tabular}{lccccr}
\toprule
Model & I & A & W & G &Error \\
\midrule
TBP (ours)    & 3 & 2& 1& 4& 50.43\%  \\
TBP (ours)   & 3 & 4& 1& 4& 46.7\%  \\
DoReFa (\citeauthor{zhou2016dorefa}) & 1 & 2& 1& 6& 53.9\%  \\
DoReFa (\citeauthor{zhou2016dorefa}) & 1 & 4& 1& 6& 51.8\%  \\
WRPN (\citeauthor{mishra2017wrpn}) & 2 & 1& 1& 32& 48.3\%  \\
WRPN (\citeauthor{mishra2017wrpn}) & 2 & 2& 2& 32& 55.8\%  \\
\hline\hline
\multicolumn{5}{c}{Single Precision -No quantiztion}\\
\hline\hline
AlexNet (\citeauthor{krizhevsky2014one}) & 1 & 32& 32& 32& 43.5\%  \tabularnewline
\bottomrule
\end{tabular}

\end{sc}
\end{small}
\end{center}
\end{table}

\begin{table}[ht]
\caption{Classification test error rates of  TBP BNNs trained on CIFAR10}
\label{simple-table}
\vskip 0.15in
\begin{center}
\begin{small}
\begin{sc}
\begin{tabular}{lcccr}
\toprule

Model&  Error Rate\\
\midrule
\multicolumn{5}{c}{Binarized activations,weights and tern gradients}
\tabularnewline
\hline 
\hline 
TBP  BNN, inflated by 3             & 9.53\%  & \tabularnewline
TBP  ResNet-18, inflated by 5       & 10.8\%  & \tabularnewline
TBP  ResNet-18, inflated by 3       & 14.21\%  & \tabularnewline
TBP  ResNet-18 \citep{he2016deep} & 18.5\%  & \tabularnewline
\hline 
\hline
\multicolumn{5}{c}{Binarized activations,weights}\tabularnewline
\hline 
\hline 
BNN \cite{hubara2016binarized}     & 10.15\%  & \tabularnewline
Binarized ResNet-18, inflated by 5   & 10.7\%  & \tabularnewline
\multicolumn{5}{c}{No binarization (standard results)}\tabularnewline
\hline 
\hline 
BNN \cite{hubara2016binarized}     & 10.94\% & \tabularnewline
ResNet-18                          & 9.64\% & \tabularnewline
\hline 
\bottomrule
\end{tabular}
\end{sc}
\end{small}
\end{center}
\vskip -0.1in
\end{table}

\subsection{Additional experiments}
To shed light on why TBP works, and on what is yet to be solved, we conducted additional set of experiments on the CIFAR-10 dataset with both ResNet-18 and the BNN like typologies.
The Back Propagation algorithm includes three phases. The forward phase in which we calculate the activations values for each layer. The Backward-Propagation (BP) phase in which the gradients from layer $\ell$ pass to layer $\ell-1$ and the update phase in which the model parameters (\emph{i.e.,} weights) are updated. Both stages require MAC operations as detailed in section 4. In this paper we focused on the BP stage. As oppose to the update stage that can be done in parallel, BP is a sequential stage. However, if the update phase uses full precision MAC operation the hardware need to support it. Moreover compressing the update gradients reduce the communication bandwidth in distributed systems. Thus, quantizing the weights gradients for the updates phase can also reduce power consummation and accelerate the training.  
% \dnote{you should clarify (here and before) what you mean by the precision of the updates and the gradients, why is it important, why did we do so far (only gradients), why is this reasonable, but also why should we try to improve and also quantize the updates. Without this explanation, this section is very unclear, and it is not clear why do we such low results in table 2, where until now the results were much better.}
\paragraph{Ternarizing both stages.}
%\dnote{what tensor? this is the first time you mention tensors in the paper, and so it is not clear what are you talking about.} 
Ternarizing both stages results with completely MAC free training. However, our results show that without enabling at least 3bit precision for the update stage the model reaches only approximately 80\% accuracy. This indicates that the ternarization noise is too high, and thus distorts the update gradients direction. If we stop the gradients ternarization once the accuracy ceases to increase, the convergence continues and the accuracy increases to the same accuracy as TBP. Thus, ternarizing the update stage can be used to accelerate TBP training of BNN networks by first training it with ternarized weights gradients and  then, for the last couple of epochs, continue training with full precision weights gradients.
%\dnote{sentence not clear.}. 
%\dnote{sentence not clear. Also, what is "update angle"?}.
%Itay: Is that clearer, feel free to change or delete this sentence
\paragraph{Multiple stochastic ternarization sampling.}
To alleviate the need for float MAC operation in the update phase we suggest to use gated XNOR operation multiple times, each time with different stochastic sample of the ternarized tensor and average the results. The algorithm is detailed in Algorithm (\ref{algoritm:multi}) and results are given in Table (\ref{tab:multi}). As expected the accuracy improves with the amount of sampling. To find the number of samples needed for each layer we adopted  a similar  geometrical approach as suggested by \citet{anderson2017high} and measured the correlation coefficient ($R$) between the update gradients received with and without ternarization. Our experiments indicate that  more samples are required for the first two convolution layer (12 samples) while the rest of the layers need approximately 6 samples. Using this configuration keeps the correlation coefficient above 0.7 and results with 87.5\% accuracy. 

\begin{algorithm}
\begin{algorithmic}
    \REQUIRE gradients from previous layer $g_{s_k}$, binarized activation $a_{k-1}^{b}$  and number of samples $S$.
    \FOR{$k=1$ to $S$} 
        \STATE $g_{s_k}^{b} \leftarrow {\rm StcTern}(g_{s_k})$
        \STATE $g_{W_k^b} \mathrel{+}= {g_{s_k}^b} a_{k-1}^{b}$
    \ENDFOR \\
    \STATE \bf{Return} $g_{W_k^b} \mathrel{/}= S$
\caption{Multiple stochastic ternarization sampling algorithm.}\label{algoritm:multi}   
\end{algorithmic}
\end{algorithm}

\begin{table}[t]
\vskip 0.15in
\begin{center}
\begin{small}
\begin{sc}
\begin{tabular}{lcccr}
\toprule
Data set & Error\\
\midrule
\hline 
TBP  BNN, 1 sample             & 20.1\%  & \tabularnewline
TBP  BNN, 5 samples            & 13.5\%  & \tabularnewline
TBP  BNN, 10 samples           & 13\%    & \tabularnewline
TBP  BNN, 20 samples           & 12\%  & \tabularnewline
TBP  BNN, $R>0.7$              & 12.5\%  & \tabularnewline
ResNet-18 10 samples    & 14\%  & \tabularnewline
ResNet-18 20 samples    & 12.7\%  & \tabularnewline
\hline 
\bottomrule
\end{tabular}
\end{sc}
\vspace{3mm}
\end{small}
 \caption{Classification test error rates of  TBP BNNs trained on CIFAR10 with Multiple Stochastic ternarization sampling for the update phase. ResNet-18 and BNN models were inflated by 5 and 3 respectively.\dnote{move table up, so it won't be stuck at the middle of the references...}}  \label{tab:multi}
\end{center}
\vskip -0.1in
\end{table}

%\subsection{Selecting the optimal quantization step %$\Delta$}
%
%In this subsection we propose a quantization method  to %compute the optimal quantization step $\Delta$ for $k$ %bit representation. Given a tensor of weights $W$ in the %following we show that $\Delta_{opt} = %\dfrac{2^k\cdot\max (W)}{N+2^{2k}}$. 
%
%The mean-square-error (MSE) between the quantified and %full precision tensor can be decomposed into variance  %and bias squared as follows [ref]:
%
%
%\begin{equation}\label{eq:50}
%\text{MSE }(W - Q(W)) = \dfrac{1}{N}} ||W - Q(W)||^2 = %\text{mean}(W - Q(W))^2 + \text{var}(W - Q(W))
%\end{equation}
%
%Our goal would be to minimize the expected value of MSE. %We assume 
%
%We assume for the sake of simplicty that we have a %single outlier 

\bibliography{nips_2018}
\bibliographystyle{nips-no-url}